\documentclass[runningheads]{llncs}
\usepackage[T1]{fontenc}
\usepackage{amsmath}
\usepackage{amsfonts}
\usepackage{amssymb}
\usepackage{subcaption} 
\usepackage{comment}
\usepackage{graphicx}
\usepackage{comment}
\usepackage{multirow}
\usepackage{url}
\usepackage{cite}
\begin{document}
%
\title{From high-frequency sensors to noon reports: Using
transfer learning for shaft power prediction in maritime}
%
%

\author{Akriti Sharma\inst{1}\orcidID{0000-0003-4623-7938} \and
Dogan Altan\inst{1}\orcidID{0000-0002-5053-4954} \and
Dusica Marijan\inst{1}\orcidID{0000-0001-9345-5431} \and
Arnbjørn Maressa\inst{2}}
\authorrunning{A. Sharma et al.}
%
\institute{Simula Research Laboratory, Kristian Augusts gate 23, 0164 Oslo, Norway \url{https://www.simula.no/} \\
\email{akriti@simula.no}
\and
Navtor AS, Copenhagen, Denmark
\url{https://www.navtor.com/}}

\maketitle              
\begin{abstract}
With the growth of global maritime transportation, energy optimization has become crucial for reducing costs and ensuring operational efficiency. Shaft power is the mechanical power transmitted from the engine to the shaft and directly impacts fuel consumption, making its accurate prediction a paramount step in optimizing vessel performance. Power consumption is highly correlated with ship parameters such as speed and shaft rotation per minute, as well as weather and sea conditions. Frequent access to this operational data can improve prediction accuracy. However, obtaining high-quality sensor data is often infeasible and costly, making alternative sources such as noon reports a viable option. In this paper, we propose a transfer learning-based approach for predicting vessels' shaft power, where a model is initially trained on high-frequency data from a vessel and then fine-tuned with low-frequency daily noon reports from other vessels.
We tested our approach on sister vessels (identical dimensions and configurations), a similar vessel (slightly larger with a different engine), and a different vessel (distinct dimensions and configurations). The experiments showed that the mean absolute percentage error decreased by 10.6\% for sister vessels, 3.6\% for a similar vessel, and 5.3\% for a different vessel, compared to the model trained solely on noon report data.

\end{abstract}

\section{Introduction} \label{sec:intro}
Efficient energy optimization for benchmarking the fuel consumption of a vessel has become essential in maritime to decrease costs and gas emissions \cite{Gkerekos2019}. The International Maritime Organization has demanded energy efficiency measures for vessels, aiming to reduce emissions from 2008 levels by 40\% in the year 2030 and 70\% by 2050 \cite{IMOsWork}. Consequently, accurately predicting power consumption is actively pursued in the maritime industry for energy efficiency planning as a vessel's fuel consumption is directly linked to its shaft power requirements \cite{Parkes2022}.

Ship-scale data can provide insights into ship performance, enabling the accurate estimation of vessel power requirements. Noon reports are more easily accessible ship-scale data \cite{Zwart2023} that provide daily readings of the vessel's location, weather and sea state estimates, as well as power and fuel consumption. Noon reports are manually generated by the captain or a data logging system, typically at ``noon" each day. The literature shows that noon reports have been used in modeling power and fuel consumption \cite{BalBeiki2016,Adland2018,Du2019}. However, noon reports are limited in terms of accuracy, as they are averaged to a single data point representing a 24-hour operation and are prone to human error \cite{Parkes2018}. 

The vast majority of vessels nowadays are also equipped with high-frequency monitoring systems that use onboard sensors to track the vessel's sailing profile, recording data every few seconds. Sensor data have less uncertainty in comparison to noon reports \cite{Aldous2015}. Switching from sensor data to noon reports has been shown to decrease the predictive error in power predictions \cite{Pedersen2009}. A comparison of noon reports and sensor data for predicting fuel consumption shows that sensor data improves model fit performance. \cite{Gkerekos2019}. However, not all vessels are equipped with sensors, and the cost of installing them is high \cite{Parkes2022}. Estimating ship operations with limited or no sensor data remains a major challenge, and in such cases, noon reports, though less accurate, are used for predictions.

In this study, we present an approach that improves the robustness and accuracy of shaft power prediction using noon reports by leveraging knowledge from sensor data of another vessel. We train a neural network on sensor data to predict shaft power, then use transfer learning to fine-tune it on noon reports from sister, similar, and different vessels. Sister vessels have identical dimensions and configurations; similar vessels differ slightly in size and design, while different vessels have distinct dimensions and configurations.
We used data-driven machine learning (ML) models for this task, as they are increasingly applied to shaft power prediction \cite{Petersen2011, Parkes2018, Mavroudis2025}. In contrast, traditional naval architecture approaches that use empirical formulas to estimate power consumption have shown a lack of scalability \cite{Zwart2023} and may not fully reflect real-world operating conditions \cite{Carlton2012, Mavroudis2025}.
Thus, we chose ML models to capture the underlying relationship in vessel power usage from noon reports. 

In this study, we consider ship data as temporal, training up to a specific year and predicting for the following year. This approach validated our method’s ability to estimate future power trends, with the fine-tuned model providing more accurate predictions. Moreover, using a pre-trained model significantly improved power predictions for noon reports compared to training from scratch. Our transfer learning-based approach is generic and can be applied to other tasks, such as trim optimization or fouling estimation in noon reports \cite{Zwart2023,Bayraktar2023}, without the need for large training datasets.

The main contributions can be summarized as follows.
\begin{itemize}
  \item We improved shaft power prediction in noon reports across a fleet by using transfer learning without needing additional training data.
  \item We bridged the performance gap in predicting shaft power that exists when transitioning from sensor data to noon reports.
  \item We modeled future power trends by forecasting subsequent year consumption patterns from historical training data.  
  \item To the best of our knowledge, this approach is the first to incorporate insights about vessel operations derived from high-frequency sensor data into daily noon reports.
\end{itemize}

\section{Literature Review} \label{sec:lit_review}
Vessel power and fuel consumption prediction has been intensely studied in the literature, using either traditional physics-based approaches or data-driven approaches built on ship-scale monitoring data. Traditional approaches use model scale towing tests or computational fluid dynamics (CFD) simulations to analyze vessel performance. The focus of these approaches is on obtaining a regression curve for the power-speed relationship in calm water and establish baseline performance \cite{Carlton2012}. A physics-based model including waves, wind and hull drift effect on power monitoring is achieving {$\sim$}5\% average error in steady state conditions \cite{Orihara2018}. Another traditional approach modeled resistance from waves along with hull form, draft, and trim corrections for estimating vessel’s speed-power performance in calm water \cite{Liu2020}. These approaches are limited by scaling effects and are seldom tested in complex sea states, failing to represent realistic vessel operations. Though empirical formulas are effective in calm water conditions \cite{Holtrop1984}, their accuracy drops in rough seas as the power consumption depends on wind and wave conditions \cite{Coraddu2017}. Furthermore, analyzing
ship performance in waves using traditional regression
methods is highly challenging \cite{Lakshmynarayanana2017}.

Data-driven approaches use ship-scale data with ML to model resistance from wind and waves for predicting the power consumption, providing an alternative to physics-based approaches. As evident from the literature, an artificial neural network (ANN) outperforms Gaussian process regression (GP) in predicting fuel consumption \cite{Petersen2011}. Indeed, a neural network-based method predicts fuel consumption for a tanker using noon reports with an error of 0.141 MT/h on mean consumption of 1.89 MT/h \cite{BalBeiki2016} and a neural network modeling daily fuel consumption for two container ships reports a root mean square error (RMSE) of 8.23 and 9.34 \cite{Du2019}. Using sensor data, the shaft power and fuel consumption are estimated with an average percentage error between 1-5\% \cite{Pedersen2009,Radonjic2014}. ML techniques have also resulted in predicting vessel's power with an average error in the range of 5\%\cite{Petursson2009,Soner2018,Laurie2021}. To improve prediction accuracy, researchers have investigated data fusion and transfer learning methods for estimating shaft power and fuel oil consumption, and we review several relevant studies in the following sections.

\subsection{Data Fusion Approaches in Power Estimation}
Researchers have systematically explored data fusion approaches from multiple sources, such as sensors, noon reports, AIS, and meteorological datasets. This research direction has evolved through several paradigms:

\textit{Noon Reports and Meteorological Data Fusion}:
A neural network-based method predicting power, improves accuracy by 2\% after combining noon reports and weather data \cite{Pedersen2009}. Another study introduced systematic data fusion of noon reports and meteorological data, achieving $R^2$ values of 0.74-0.90 across eight containerships \cite{Li2022}. Data from noon reports, AIS, meteorological sources, and on-board sensors were combined to develop models that summarize fuel consumption during a journey \cite{Man2020} and monitoring of engine performance \cite{Uyank2020}.

\textit{Sensor Data and Meteorological Data Fusion}: Extending the data fusion concept to high-frequency sensor data, a study modeling ship fuel efficiency reported $R^2$ value above 0.96 \cite{Du2022}.

These approaches advance data fusion but remain limited by data availability and vessel-specific constraints.
\subsection{Transfer Learning}
Transfer learning has been applied to vessel power prediction through several approaches, each addressing a different challenge.

\textit{Physics-to-Data Transfer Learning}: Transfer learning was employed for vessel power prediction using synthetic data from physics-based simulations with sensor data as the source domain and fine-tuning with real operational data, showing improved accuracy with reduced training data \cite{Mavroudis2025}. Our research focuses on data-driven knowledge transfer between sensor data and noon reports, using neural networks to intrinsically capture physical relationships among input variables.

\textit{Same-Frequency Data Transfer Learning}:
Transfer learning has been explored for fuel consumption prediction between sister vessels after using three strategies: (a) fine-tuning with freezing transferred layers, (b) fine-tuning without freezing transferred layers, and (c) optimizing network structure with freezing transferred layers \cite{Luo2025}. Their approach addressed the challenge of limited training data, a common issue for new ships with limited operating time. Results showed MAPE reductions of 12.57\%, 6.44\%, and 16.03\%, respectively. Our approach achieved a 9.6\% reduction in MAPE for sister vessels, but a direct comparison will be inappropriate as we use transfer learning between high and low-frequency data, while they used the same frequency data.

\textit{Fleet-Based Prediction Without Transfer Learning}: Another shaft power modeling approach trained neural networks on aggregated fleet data without transfer learning, demonstrating accurate power prediction (1.5–5\% error) using high-frequency data despite limited records \cite{Parkes2022}. Our approach enables cross-fleet but with cross-frequency adaptation rather than aggregated fleet data.

Despite advances in data fusion and transfer learning, including same-frequency transfer learning \cite{Luo2025}, fleet-wide fusion within similar data types \cite{Parkes2022}, and fusion within similar temporal domains \cite{Li2022, Du2022}, no work has addressed knowledge transfer across different data resolutions (high-frequency sensor data to low-frequency noon reports). Existing approaches focus on sister vessels \cite{Luo2025} or require fleet-wide data from similar vessels \cite{Parkes2022}, but evaluations across diverse vessel types and operations remain unexplored. High-performance approaches demand specialized infrastructure (sensor installations, meteorological data access), large-scale fleet data, creating barriers for resource-limited operators. Additionally, these approaches rely on random data splits instead of temporal validation, limiting a real-world deployment, where models must forecast future performance based on historical data.

This study addresses these gaps by proposing the first cross-frequency transfer learning framework, transferring knowledge from high-frequency sensor data to low-frequency noon reports. We evaluate its effectiveness across different vessel types and and validate it through temporal forecasting, without requiring additional sensor datasets, or fleet-wide data collection. Table~\ref{tab0} provides a  detailed description of the research context and shows how this work addresses a specific gap in maritime transfer learning.
\begin{table}[htbp]
\caption{Research contribution context: our study addressing relevant gap in maritime transfer learning. (NRs: Noon Reports)}
\begin{tabular}{|l|l|l|}
\hline
\textbf{Research Direction} & \textbf{Addressed Challenge} & \textbf{Our Contribution} \\ \hline
Data Scarcity & Only high frequency sensor data & Cross-frequency transfer \\ \hline
Vessel Similarity & Sister vessels analysis & Multi-vessel analysis \\ \hline
Deployment & Random splits (laboratory tested) & Temporal forecasting tested \\ \hline
Practical Adoption & Expensive \& complex setup needed & Easily accessible NRs \\ \hline
\end{tabular}
\label{tab0}
\end{table}

\section{Data} \label{sec:data}
The dataset used in this study was collected from seven vessels: five general cargo ships and two container ships. The data belongs to our industrial partner and, due to confidentiality, it is not openly available. To maintain data anonymity, each vessel is represented using its type acronym and numeric code. For the cargo category, four sister vessels are denoted as S\_V followed by their number and a similar vessel as SM\_V1. The containers are referred to as different vessels, D\_V1 and D\_V2. Sister vessels have identical dimensions (length 204 and beam 32 meters) and the same configuration. The similar vessel is slightly longer and broader than the sister vessels and has a different main engine configuration. The different vessels' dimensions (length 209 and beam 30 meters) and configurations are different from those of sister vessels. 

Sensor data reports shaft power at 15-minute intervals, and noon reports are recorded once every 24 hours. The study is performed on a per-vessel basis, and data for each vessel is temporally split into train and test sets. For sister and similar vessels, the train set is up to the year 2023, and the test set is from 2024. For the other vessels, the train set is for 2024 and the test set is for 2025. The yearly distribution varies due to data availability. We aim to keep the experiment realistic, as vessels often have historical data from different periods, and testing across years will validate if our approach captures trends in noon reports. Table~\ref{tab1} provides a more detailed description of the dataset, with each column showing the vessel name, dataset time-span, and the number of instances in the training and test sets.

\begin{table}[htbp]
\caption{Data instances for each vessel's sensor data (SD) and noon reports (NR), along with their corresponding time spans.}
\begin{tabular}{|c|c|c|c|c|c|}
\hline
\textbf{Vessel} & \textbf{Timespan} & \multicolumn{1}{l|}{\textbf{SD (Train)}} & \multicolumn{1}{l|}{\textbf{SD (Test)}} & \textbf{NR (Train)} & \textbf{NR (Test)} \\ \hline
S\_V1 & Jan'23-Aug'24 & 14661 & 13040 & 681 & 138 \\ \hline
S\_V2 & May'20-Aug'24 & 26948 & 23575 & 783 & 145 \\ \hline
S\_V3 & May'20-Aug'24 & 31047 & 11739 & 846 & 137 \\ \hline
S\_V4 & April'22-Aug'24 & 22133 & 287 & 401 & 135 \\ \hline
SM\_V1 & May'20-Aug'24 & 25592 & 6659 & 725 & 73 \\ \hline
D\_V1 & April'24-May'25 & 26948 & 18596 & 208 & 101 \\ \hline
\end{tabular}
\label{tab1}
\end{table}

\subsection{Feature Selection}
Noon reports for sister vessels have 41 features, and different vessels have 65, out of which we selected seven representative features for predicting the target feature, shaft power (kilowatts). The features are: speed through water (knots), shaft rotational speed (revolution per minute (RPM)), draft amidships (meters), wave height (meters), swell height (meters), wave direction (degrees) and wind direction (degrees). Wave and wind directions are relative to the vessel; wave height represents the significant wave height, while swell height represents the significant height of the wave component caused by wind. As evident in the literature, using wave and wind data improves power prediction accuracy \cite{Parkes2020, Parkes2022}, and swell height is significant for fuel consumption predictions from noon reports \cite{Aldous2013}. We select speed through water over speed over ground since current data is captured by the wave component. The input feature selection also uses on Pearson Correlation analysis with shaft power to identify relevant variables:
$$r_{x_i,y} = \frac{\sum_{j=1}^{n}(x_{ij} - \bar{x}_i)(y_j - \bar{y})}{\sqrt{\sum_{j=1}^{n}(x_{ij} - \bar{x}_i)^2}\sqrt{\sum_{j=1}^{n}(y_j - \bar{y})^2}}$$
where $x_i$ represents the i-th candidate feature, y represents shaft power, and n is the number of observations.
The final feature vector is defined as:
$$\mathbf{x} = [v_{stw}, \omega_{rpm}, d_{mid}, H_{wave}, H_{swell}, \phi_{wave}, \phi_{wind}]^T \in \mathbb{R}^7$$
where each component represents:
\begin{itemize}
  \item[] $v_{stw}, \omega_{rpm}$: speed through water and RPM
  \item[] $d_{mid}$: draft amidship [$(draft_{aft}+ draft_{fore})/2$]
  \item[] $H_{wave}, H_{swell}$: wave and swell heights
   \item[] $\phi_{wave}, \phi_{wind}$: wave and wind directions (relative to vessel)
\end{itemize}

The same features are used for sensor data to enable transfer learning with noon reports. The sensor data had high uncertainty regarding weather conditions, which can be resolved by the use of publicly accessible meteorological data \cite{Man2020,Li2022,Du2022}. We derived wave, wind, and swell components from the Copernicus Marine Service (CMEMS) databases, with 0.083° spatial and 3-hour temporal resolution. AIS data enabled spatial and temporal interpolation between CMEMS observations using trilinear interpolation. The weather data fusion from CMEMS databases is expressed as: 
$$x_{weather}(time, lat, lon) = \text{TriLinear}(\mathbf{X}_{CMEMS}, time, lat, lon)$$
where $TriLinear$  represents trilinear interpolation over spatial and temporal coordinates. The CMEMS direction data was relative to north, so we converted it to be relative to the ship's course:
$$\phi_{relative} = (\phi_{CMEMS} - AIS Course_{vessel}) \% 360$$

\section{Methodology} \label{sec:method}
In this section, we provide background information on transfer learning and introduce our transfer learning-based method for predicting the shaft power. 
\subsection{Transfer Learning}
Transfer learning is a technique where a pre-trained model on a source task or dataset is fine-tuned for a different task or dataset \cite{Weiss2016}. It uses the weights from the source task's network layers to partially retrain the model for the target task. A pre-trained model is a network trained on a benchmark dataset to solve a task. This pre-trained network can be retrained using the same or different target task-specific data by fine-tuning one or more of its layers. Fine-tuning requires pre-trained weights to partially or fully retrain the network for the target task, allowing it to learn features specific to both the source and target tasks. In this study, both the source and target tasks are shaft power prediction. The proposed approach utilizes a pre-trained model for power consumption from sensor data, following transfer learning principles, we define the source domain as $\mathcal{D}_S = \{(\mathbf{x}_i^s, y_i^s) | i = 1, \ldots, n_s\}$ representing sensor data, and target domain as
$\mathcal{D}_T = \{(\mathbf{x}_j^t, y_j^t) | j = 1, \ldots, n_t\}$ representing noon reports. The transfer is formulated as:
$$\theta_{target} = \arg\min_{\theta} \mathcal{L}(f(\mathbf{x}_{noon}; \theta_{frozen}, \theta_{tune}), y_{noon})$$
where $\theta_{frozen}$ represents pre-trained parameters from sensor data (layers 1-3), $\theta_{target}$ represents parameters adapted for noon reports (layer 4) and $\mathcal{L}$ is the loss function.

\subsection{Proposed Method}
Our approach applies transfer learning to a neural network-based regression model trained on high-frequency sensor data, fine-tuning it to predict shaft power using the low-frequency data found in noon reports. The architecture of our model is inspired by a study predicting shaft power for vessels across sister vessels \cite{Parkes2022}. After extensive experimentation, which included adding extra hidden layers, adjusting the number of neurons, adding/removing dropout layers, and evaluating the impact of train and test across the entire dataset, we selected a (128, 64, 32, 1) configuration with no drop out layer for our model. The input dimension is seven, and each layer, except the final one, uses ReLU activation. Mean absolute error (MAE) is used to calculate the loss, similar to another transfer learning-based framework predicting fuel consumption \cite{Luo2025} and Adam optimizer is used. The approach considers the model as the baseline when training or evaluating sensor data for shaft power prediction. The baseline neural network implements a feedforward architecture:
$$f(\mathbf{x}; \theta) = W^{(4)} \cdot \sigma(W^{(3)} \cdot \sigma(W^{(2)} \cdot \sigma(W^{(1)} x + b^{(1)}) + b^{(2)}) + b^{(3)}) + b^{(4)}$$
where $W^{(i)} \in \mathbb{R}^{n_i \times n_{i-1}}$ are weight matrices and $b^{(i)} \in \mathbb{R}^{n_i}$ are bias vectors for layer $i$, with network dimensions (7,128,64,32,1), ReLU activation function $\sigma(z) = max(0,z)$.

The baseline model optimizes mean absolute error loss:
$$\mathcal{L}_{baseline} = \frac{1}{n} \sum_{i=1}^{n} |y_i - \hat{y}_i|$$
where $y_i$ is the actual shaft power and $\hat{y}_i$ is the predicted value.The transfer learning model when fine-tuned on noon reports combines frozen feature extraction with gradient computation having selective parameters update:
$$\mathcal{L}_{final} = \frac{1}{n_t} \sum_{i=1}^{n_t} |y_i^{noon} - f(x_i^{noon}; \theta_{frozen}, \theta_{tune})|$$
The final model architecture will be:
\begin{align*}
f_{final}(x) = W_{new}^{(4)} \cdot \sigma(W_{frozen}^{(3)} \cdot \sigma(W_{frozen}^{(2)} \cdot \sigma(W_{frozen}^{(1)} x + b_{frozen}^{(1)}) +b_{frozen}^{(2)}) \\
+ b_{frozen}^{(3)}) + b_{new}^{(4)}
\end{align*}

\subsubsection{Baseline Model}
The regression model is trained on the training set of vessels' sensor data for 300 epochs with a batch size of 32 and a learning rate of $10^{-3}$. The model uses a \textit{ReduceLROnPlateau} scheduler, reducing the learning rate by 0.5 after three epochs if the validation loss does not improve, and employs an early stopping scheme with a patience of five epochs. The weights with the minimum validation loss are saved for the baseline model.

\subsubsection{Final Model}
The pre-trained regression model (baseline) is fine-tuned on the train set of noon reports for vessels S\_V2, S\_V3, S\_V4, SM\_V1, D\_V1 and D\_V2. All layers except the last are frozen during training. The training configuration is identical to the baseline model, except for the batch size and learning rate, which are set to 16 and $10^{-4}$ respectively.

\begin{figure}
\includegraphics[width=\columnwidth]{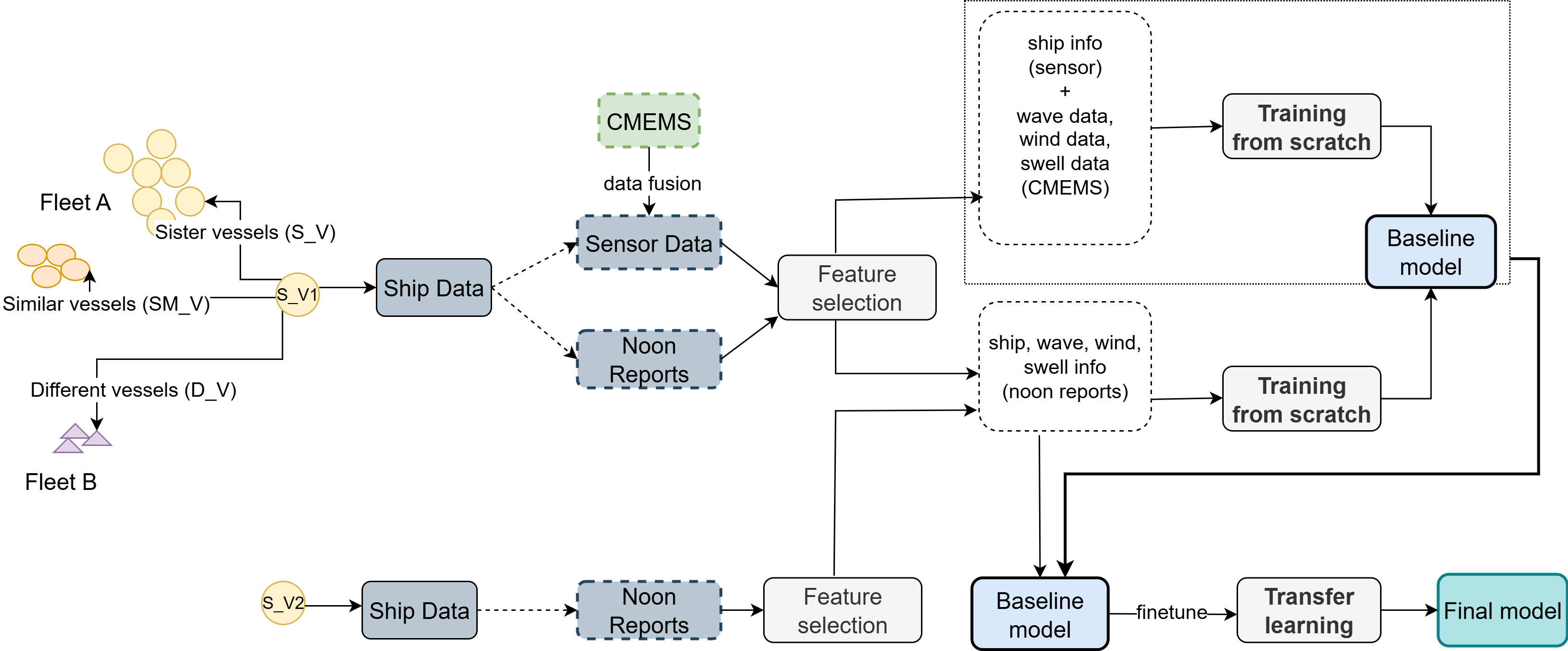}
\caption{Overview of the proposed shaft power prediction method. Baseline models are trained using ship data from vessel S\_V1, which includes noon reports and sensor data. The sensor data is fused with Copernicus (CMEMS) data. The baseline model trained on sensor data is fine-tuned using noon reports from sister vessel S\_V2 through our transfer learning approach, resulting in the final model predicting shaft power.}
\label{method-overview}
\end{figure}

Figure ~\ref{method-overview} shows all the above-described steps of the proposed transfer learning-based shaft power prediction method.

\subsubsection{Data Preprocessing}
The data was preprocessed to remove data instances with a speed through water of less than 2 knots, RPM of less than 1, and reported power of less than 1 kilowatt. We observed outliers in noon reports exceeding 12,000 kilowatts and removed them accordingly.

\section{Experimental Setup}  \label{sec:exp_setup}
We train the baseline model using the train set of sensor data from sister vessels, as well as similar and different vessels. The vessel achieving the best predictive performance on the test set is selected as the baseline model for our proposed approach. The train set of noon reports from the remaining vessels in the sister, similar, and different categories is then used for fine-tuning the baseline model on a per-vessel basis.
Furthermore, we train a separate baseline model from scratch using only the noon reports from the vessels. This enables a comparative analysis of the predictive performance between the models based on sensor data while also benchmarking the improvements in accuracy and robustness of our proposed method in predicting shaft power from noon reports. To mitigate the initialization bias or the data sampling bias, we ran the experiments 10 times and reported the mean and standard deviation (SD) of the evaluation metrics. We analyzed the evaluation results based on these research questions:

\begin{itemize}
  \item[] \textbf{RQ1} What is the performance decline in shaft power prediction when using noon reports instead of sensor data?
  \item[] \textbf{RQ2} Does predictive performance for shaft power improve with the proposed method?
  \item[] \textbf{RQ3} Can the trend of improved predictive performance be observed when the model is trained on one year of data and tested on the following year's data?
\end{itemize}

\subsection{Evaluation metrics}
We evaluate the performance of our experiments with the following metrics:
\begin{itemize}
    \item Mean absolute error (MAE): MAE is a performance metric that evaluates regression models by measuring the mean absolute difference between the labels and predicted values of a model. 
\begin{equation}
\text{MAE} = \frac{1}{n} \sum_{i=1}^{n} \left| y_i - \hat{y}_i \right|
\end{equation}

  where $y_i$ is the actual value (label), $\hat{y}_i$ is the predicted value and $n$ is the number of data instances. 
  
  \item Normalized mean absolute error (NMAE): MAE does not consider the scale of the target labels, which can vary across different datasets. Normalizing the MAE by the mean of the target values provides a better understanding of the model's performance across different data distributions, such as sensor data versus noon reports. NMAE is calculated using the following formula.
\begin{equation}
\text{NMAE} = \frac{\text{MAE}}{\text{max}(y) - \text{min}(y)}
\end{equation}
  
    \item Mean absolute percentage error (MAPE): MAPE presents the error in prediction as a percentage of actual values and is calculated as follows:
\begin{equation}
\text{MAPE} = \frac{100\%}{n} \sum_{i=1}^{n} \left| \frac{y_i - \hat{y}_i}{y_i} \right|
\end{equation}

    \item Coefficient of determination (R$^2$): $R^2$ is a performance metric that measures how well the predictions of the model approximate the actual data. The following formula is used for calculation:   
\begin{equation}
R^2 = 1 - \frac{\sum_{i=1}^{n} (y_i - \hat{y}_i)^2}{\sum_{i=1}^{n} (y_i - \bar{y})^2}
\end{equation}

where $\bar{y}$ is the mean of actual values (labels).
\end{itemize}

\section{Results and Analysis} \label{sec:results}
In this section, we present our experimental results, where the model is trained on both sensor data and the noon reports training set to compare its predictive performance on shaft power. The model trained with sensor data serves as the baseline and is fine-tuned to the final model using noon reports, based on our transfer learning-based shaft power prediction approach. Predictive accuracy is evaluated for the final model and compared to models trained from scratch using both noon reports and sensor data.
Finally, vessel power consumption for 2024 and 2025 is predicted using models trained with our method and from scratch. This helps analyze the robustness of the forecasted consumption trend across the entire noon report test set.

\subsection{Comparison of vessel power prediction: sensor data vs. noon reports}
We evaluate the performance of models trained on sensor data and noon reports. We conduct this experiment to observe a decline in shaft power performance when switching between these two data sources, answering RQ1.

Table~\ref{tab2} presents the evaluation metrics for shaft power prediction of vessels, using the test set for both sensor data and noon reports. The reported R$^2$ was higher and MAPE was lower for sensor data compared to noon reports for all vessels. To specifically answer RQ1, we use NMAE for direct comparison. An average error increase of 4\% (0.04) was observed for sister vessels, 7\% (0.07) for similar vessels, and 3\% (0.03) for different vessels when transitioning from sensor data to noon reports. Also, for this experiment,
the reported SD for MAPE was less than 2, and NMAE and R$^2$ were less than 1 for all vessels. Since S\_V1 outperformed the other vessels on both datasets, we selected it as the base model for our approach, which will be evaluated in the subsequent experiments.

\begin{table}
\caption{Shaft power predictions using sensor data and noon reports.}
\begin{tabular}{|c|ccc|ccc|}
\hline
\multirow{2}{*}{\textbf{Vessel}} & \multicolumn{3}{c|}{\textbf{Sensor Data}} & \multicolumn{3}{c|}{\textbf{Noon Reports}} \\ \cline{2-7} 
 & \multicolumn{1}{c|}{\textbf{R$^2$}} & \multicolumn{1}{c|}{\textbf{NMAE}} & \textbf{MAPE} & \multicolumn{1}{c|}{\textbf{R$^2$}} & \multicolumn{1}{c|}{\textbf{NMAE}} & \textbf{MAPE} \\ \hline
S\_V1 & \multicolumn{1}{c|}{0.97} & \multicolumn{1}{c|}{0.02} & 3.54 & \multicolumn{1}{c|}{0.77} & \multicolumn{1}{c|}{0.08} & 11.19 \\ \hline
S\_V2 & \multicolumn{1}{c|}{0.81} & \multicolumn{1}{c|}{0.06} & 16.18 & \multicolumn{1}{c|}{0.37} & \multicolumn{1}{c|}{0.10} & 54.00 \\ \hline
S\_V3 & \multicolumn{1}{c|}{0.44} & \multicolumn{1}{c|}{0.09} & 28.05 & \multicolumn{1}{c|}{-0.32} & \multicolumn{1}{c|}{0.13} & 52.00 \\ \hline
S\_V4 & \multicolumn{1}{c|}{0.54} & \multicolumn{1}{c|}{0.10} & 5.00 & \multicolumn{1}{c|}{0.06} & \multicolumn{1}{c|}{0.15} & 22.00\\ \hline
SM\_V1 & \multicolumn{1}{c|}{0.80} & \multicolumn{1}{c|}{0.05} & 8.59 & \multicolumn{1}{c|}{0.19} & \multicolumn{1}{c|}{0.12} & 18.00\\ \hline
D\_V1 & \multicolumn{1}{c|}{0.95} & \multicolumn{1}{c|}{0.03} & 0.13 & \multicolumn{1}{c|}{0.85} & \multicolumn{1}{c|}{0.06} & 28.00 \\ \hline
\end{tabular}
\label{tab2}
\end{table}

\subsection{Improving the predictive accuracy of vessel power from noon reports}
In this experiment, we evaluated the model's performance in predicting shaft power for two settings: training from scratch and training based on our transfer learning approach. We conducted this experiment to observe if our proposed approach improves shaft power prediction using noon reports to compare its performance with predictions based on sensor data, thereby answering RQ2.

Table~\ref{tab3} presents the evaluation metrics for vessel shaft power prediction, comparing transfer learning with training from scratch on the noon reports test set. For each vessel, our transfer learning-based method outperformed the training from scratch approach in predicting shaft power. We achieved an average 10.6\% reduction in MAPE for sister vessels, 3.6\% for similar and 5.3\% for different vessels. Furthermore, the R$^2$ improved for all vessels, with a significant increase in magnitude and a sign change from positive to negative for S\_V3. We also observed a reduction in MAE; for sister vessels, it was by a factor of 200, for similar vessels by 152, and 316 for different vessels.
This clearly indicates that our approach significantly improved shaft power prediction in noon reports. 

\begin{table}
\caption{Shaft power predictions on noon reports: training from scratch vs. transfer learning (TL).}
\begin{tabular}{|c|ccc|ccc|}
\hline
\multirow{2}{*}{\textbf{Vessel}} & \multicolumn{3}{c|}{\textbf{Training from scratch}} & \multicolumn{3}{c|}{\textbf{Training based on TL}} \\ \cline{2-7} 
 & \multicolumn{1}{c|}{\textbf{R$^2$}} & \multicolumn{1}{c|}{\textbf{MAE}} & \textbf{MAPE} & \multicolumn{1}{c|}{\textbf{R$^2$}} & \multicolumn{1}{c|}{\textbf{MAE}} & \textbf{MAPE} \\ \hline
S\_V2 & \multicolumn{1}{c|}{0.37} & \multicolumn{1}{c|}{657.49} & 54.00 & \multicolumn{1}{c|}{0.76} & \multicolumn{1}{c|}{357.40} & 38.00 \\ \hline
S\_V3 & \multicolumn{1}{c|}{-0.32} & \multicolumn{1}{c|}{768.61} & 52.00 & \multicolumn{1}{c|}{0.35} & \multicolumn{1}{c|}{494.58} & 41.00 \\ \hline
S\_V4 & \multicolumn{1}{c|}{0.06} & \multicolumn{1}{c|}{988.90} & 22.00 & \multicolumn{1}{c|}{0.31} & \multicolumn{1}{c|}{767.88} & 17.21 \\ \hline
SM\_V1 & \multicolumn{1}{c|}{0.19} & \multicolumn{1}{c|}{911.78} & 18.00 & \multicolumn{1}{c|}{0.45} & \multicolumn{1}{c|}{759.38} & 14.36 \\ \hline
D\_V1 & \multicolumn{1}{c|}{0.85} & \multicolumn{1}{c|}{975.03} & 28.00 & \multicolumn{1}{c|}{0.91} & \multicolumn{1}{c|}{659.83} & 22.70 \\ \hline
\end{tabular}
\label{tab3}
\end{table}

Next, we compared the performance of the transfer learning-based shaft power prediction for a vessel with the model trained on sensor data for that same vessel.
Table~\ref{tab4} presents the evaluation metrics for power prediction using training from scratch with sensor data and noon reports, along with our proposed transfer learning approach using noon reports. The results showed that our  proposed method bridges the performance gap in terms of the model's predictive capability when transitioning from sensor data to noon reports. The previously reported 4\% increase in NMAE for this transition was reduced to an average of 1.3\% (0.013) for sister vessels. For a similar vessel, NMAE decreased from 7\% to 4\% (0.04), and for a different vessel, it improved from 3\% to 2\% (0.02). This improvement in predictive accuracy for shaft power using noon reports answers RQ2. 

For these experiments, the reported SD for MAPE was less than 2 and for NMAE and R$^2$ it was less than 1 for all datasets.

\begin{table}
\caption{Shaft power predictions on different datasets: training from scratch vs. transfer learning (TL).}
\begin{tabular}{|c|c|c|c|}
\hline
\textbf{Vessel} & \textbf{\begin{tabular}[c]{@{}c@{}}NMAE \\ (sensor data)\end{tabular}} & \textbf{\begin{tabular}[c]{@{}c@{}}NMAE \\ (noon reports with TL)\end{tabular}} & \textbf{\begin{tabular}[c]{@{}c@{}}NMAE \\ (noon reports)\end{tabular}} \\ \hline
S\_V2 & 0.06 & 0.07 & 0.10 \\ \hline
S\_V3 & 0.09 & 0.09 & 0.13 \\ \hline
S\_V4 & 0.10 & 0.13 & 0.15 \\ \hline
SM\_V1 & 0.05 & 0.09 & 0.12 \\ \hline
D\_V1 & 0.03 & 0.05 & 0.06 \\ \hline
\end{tabular}
\label{tab4}

\end{table}

\subsection{Assessing the improved accuracy of predictions by analyzing the power consumption trend}
In this experiment, we present the actual and predicted shaft power values for different settings: models trained from scratch and models trained using the proposed transfer learning approach. We conducted this experiment to observe whether our method can accurately predict power consumption for the upcoming years, 2024 and 2025, despite the base model being trained with data from 2023, thus addressing RQ3.

\begin{figure*}[htbp]
    \centering
    \begin{minipage}{\textwidth}
        \centering
        \includegraphics[width=\textwidth]{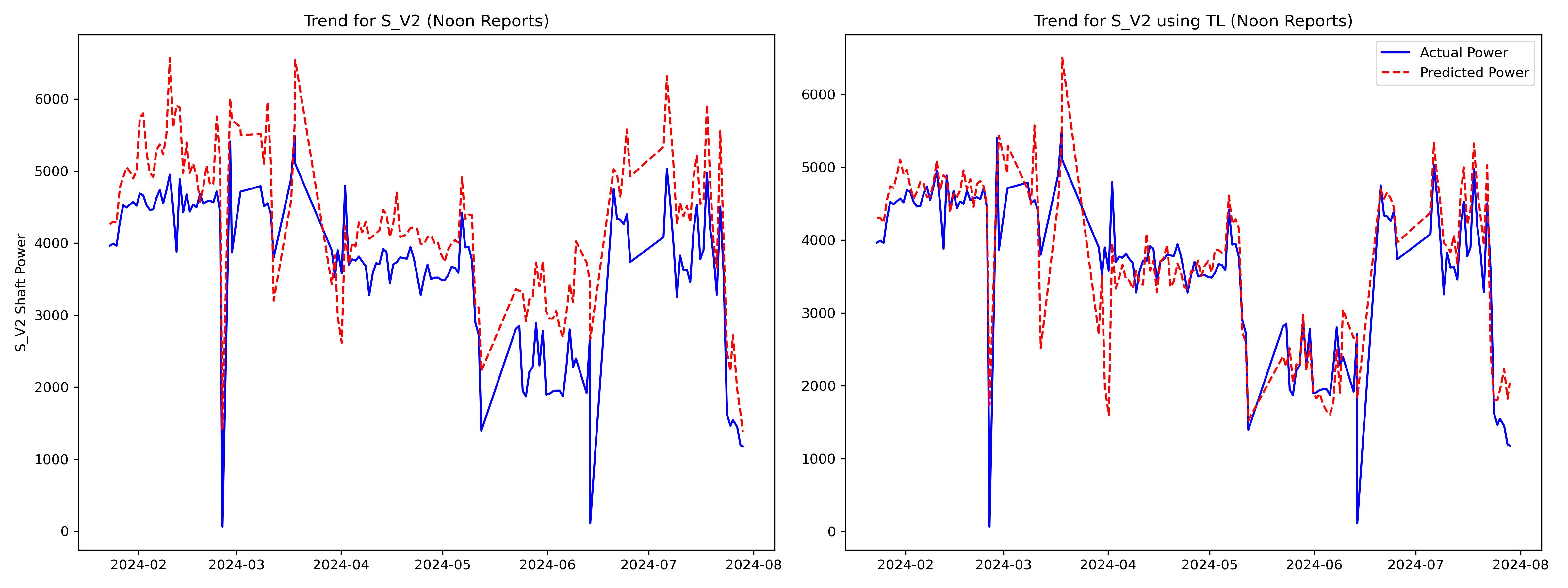}
        \subcaption{Shaft power for S\_V2 in the year 2024}
    \end{minipage}
    \vskip\baselineskip
    \begin{minipage}{\textwidth}
        \centering
        \includegraphics[width=\textwidth]{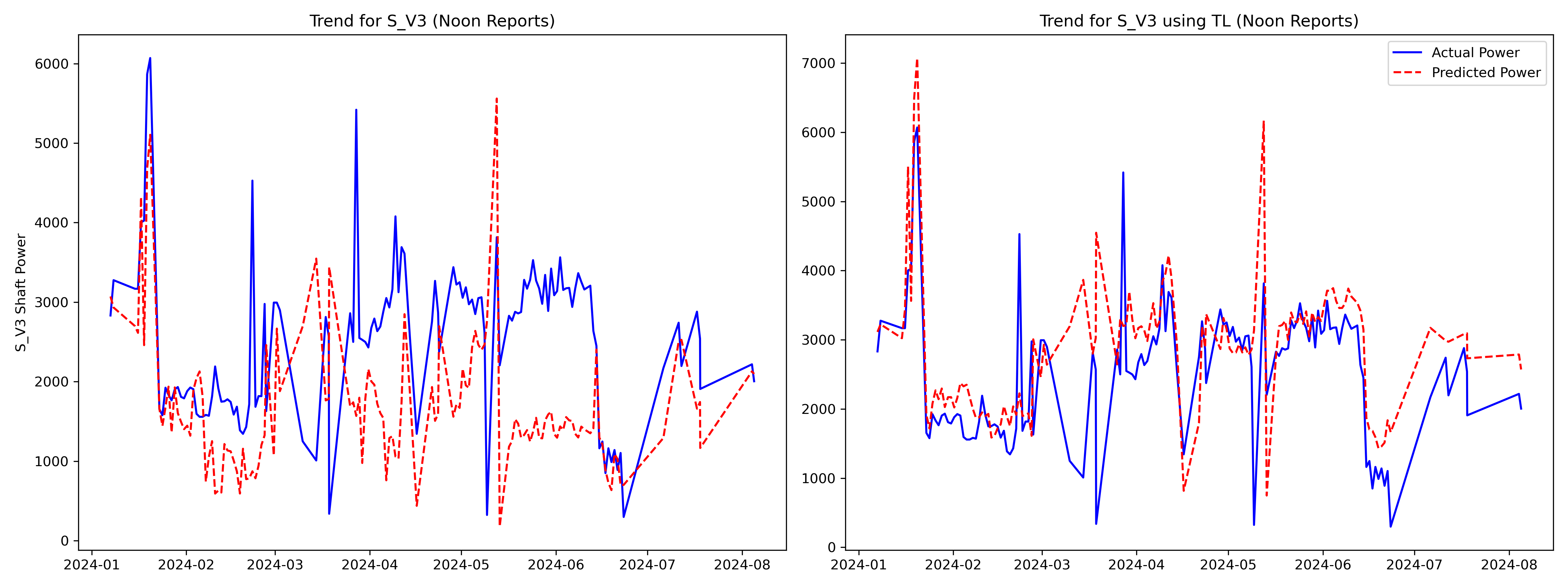}
        \subcaption{Shaft power for S\_V3 in the year 2024}
    \end{minipage}
    \vskip\baselineskip
    \begin{minipage}{\textwidth}
        \centering
        \includegraphics[width=\textwidth]{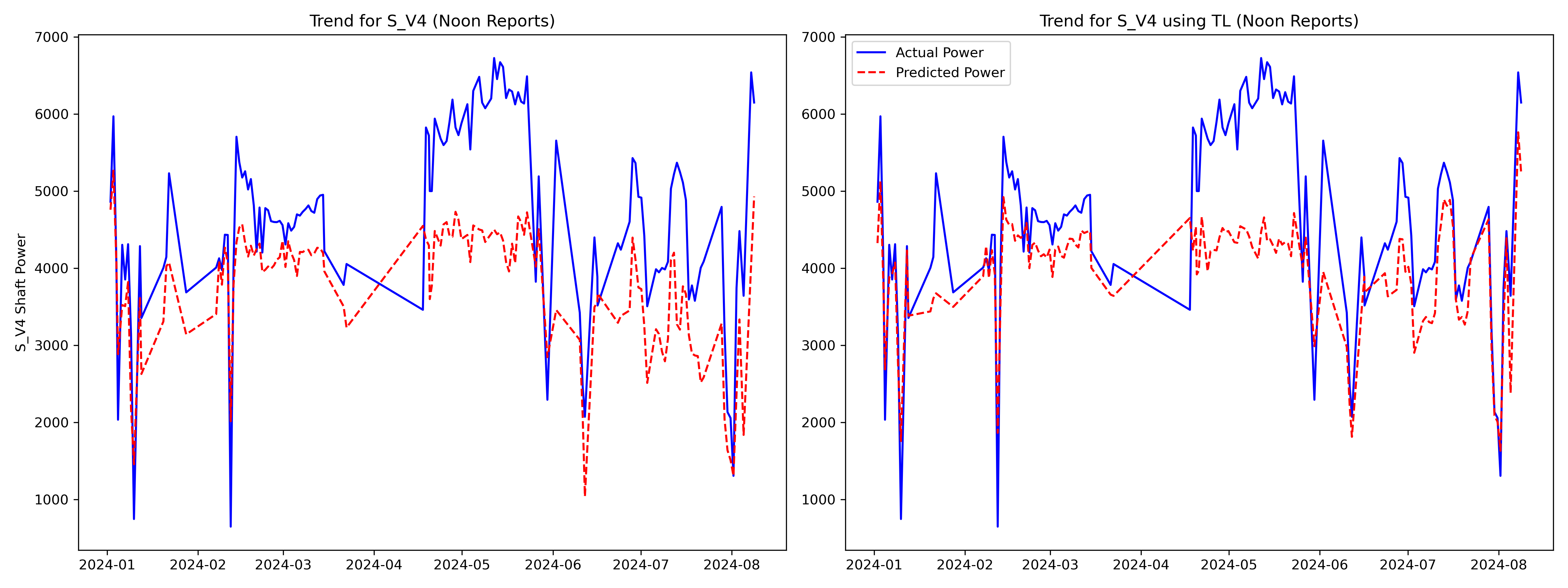}
        \subcaption{Shaft power for S\_V4 in the year 2024}
    \end{minipage}
    \caption{Sister vessels: forecasted consumption trends based on power predictions by models trained from scratch and transfer learning (TL).}
    \label{fig:group1}
\end{figure*}

Fig.~\ref{fig:group1} shows the shaft power prediction results for sister vessels using both training from scratch and transfer learning approaches on the noon reports test set. Our method provides a more accurate prediction of the power consumption trend for all sister vessels compared to training from scratch. The accuracy of the forecasted power requirements was consistent not only in the initial months of 2024 (January and February) but also towards the end (August and September). For S\_V3, although the predictive accuracy decreased around mid-2024, the trend was still better represented than the baseline model trained from scratch.

\begin{figure*}[htbp]
    \centering
    \begin{minipage}{\textwidth}
        \centering
        \includegraphics[width=\textwidth]{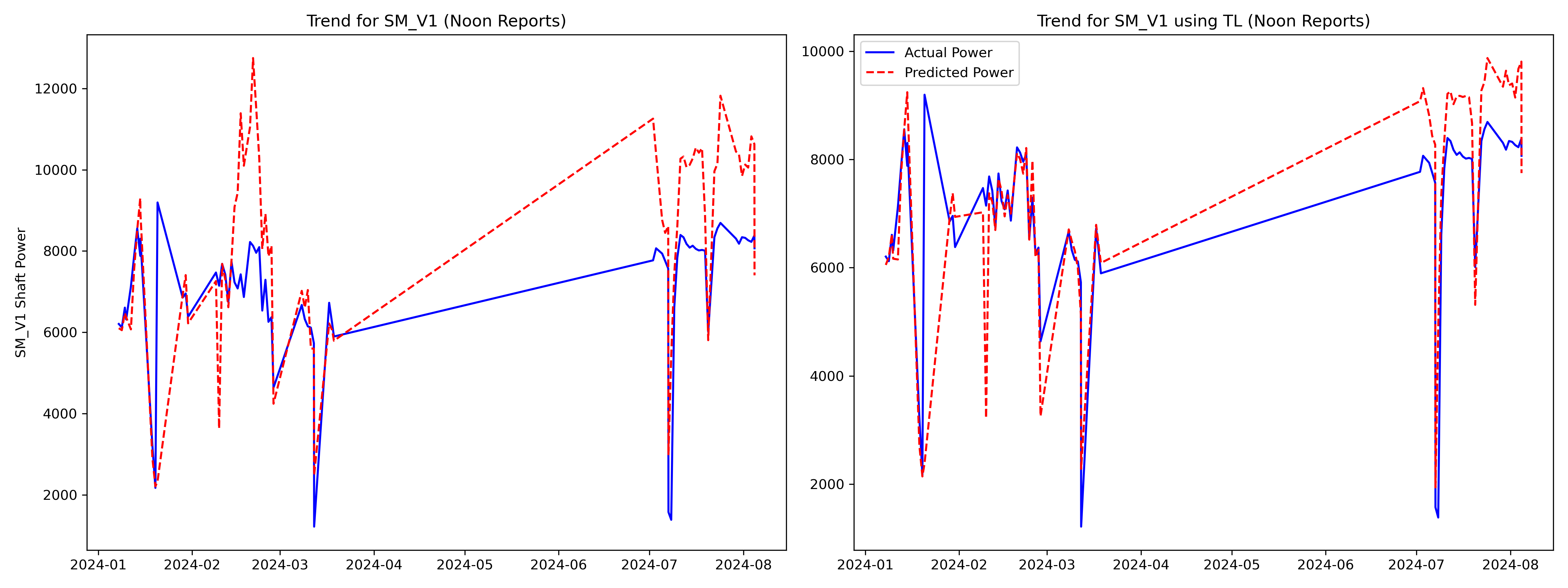}
        \subcaption{Shaft power for SM\_V1 in the year 2024}
    \end{minipage}
    \vskip\baselineskip
    \begin{minipage}{\textwidth}
        \centering
        \includegraphics[width=\textwidth]{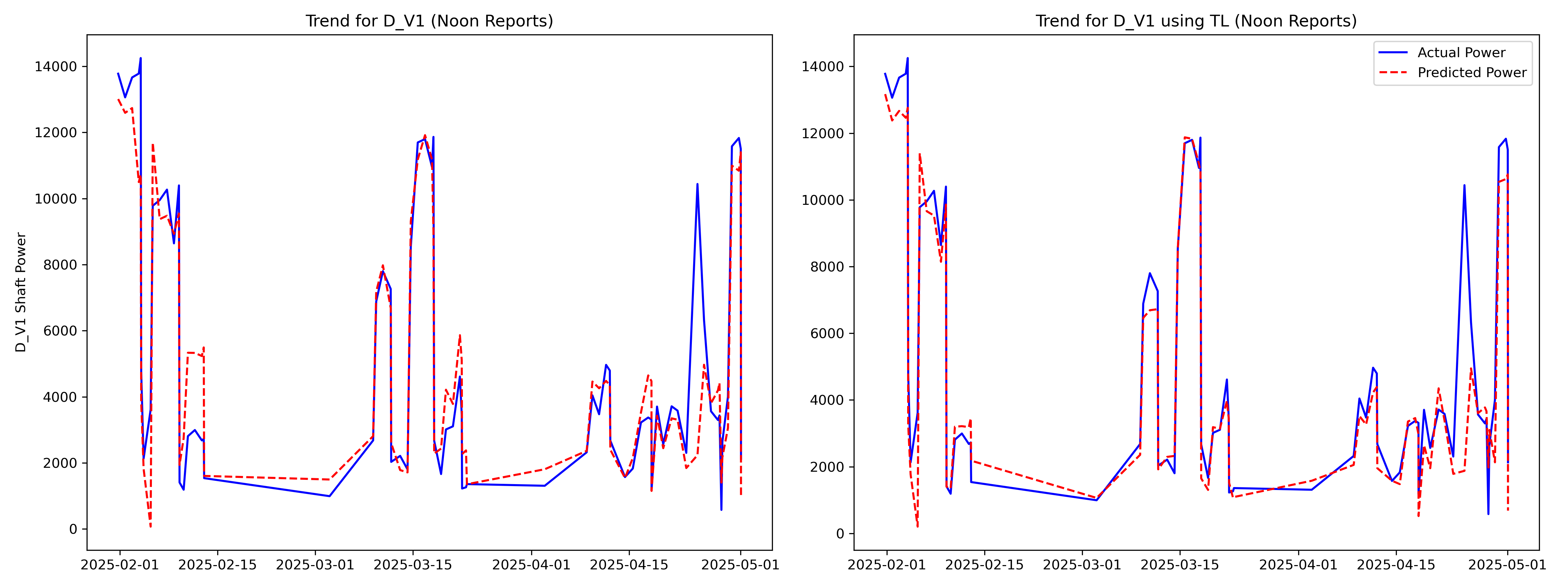}
        \subcaption{Shaft power for D\_V1 in the year 2025}
    \end{minipage}   
    \caption{Non-sister vessels: forecasted consumption trends based on power predictions by models trained from scratch and transfer learning (TL).}
    \label{fig:group2}
\end{figure*}

Fig.~\ref{fig:group2} shows the shaft power predictions for non-sister vessels (similar and different vessels) using both training from scratch and transfer learning approaches on the noon reports test set. Similar to sister vessels, our method provides more accurate predictions of consumption trends for non-sister vessels compared to training from scratch. Despite data scarcity from April to September 2024 for SM\_V1, our transfer learning approach predicted the consumption trend with higher accuracy and robustness. For D\_V1, both training approaches reported similar accuracy in predicting the power consumption trend. Although the base model (a different vessel, S\_V1) was trained on data up to 2023 and the forecast was for data collected in 2025, our approach still predicted with high accuracy. Thus, answering RQ3, the trend of improved predictive performance was observed for 2024 and 2025.

\section{Conclusion and Discussion} \label{sec:discussion}
In this paper, we present a transfer learning-based method to improve shaft power prediction for a vessel using noon reports. Our approach first trains a model on a vessel's sensor data for shaft power prediction, then fine-tunes the model using noon reports from other vessels. Data from multiple vessels, including sisters, a similar and a different vessel, are used to evaluate the method's performance.

The experimental results show that the baseline model trained on a single sister vessel, when fine-tuned with noon reports from other sisters, reduced the error by 10.6\% in predicting shaft power compared to a model trained solely on noon reports from those sisters. In comparison to sensor data models, our approach resulted in an average error increase of 1.3\%, while models trained solely on noon reports showed a 4\% increase in error. Therefore, our approach effectively reduced the performance gap in shaft power prediction when transitioning from sensor data to noon reports, with the reported performance nearly matching to sensor data. Hence, when sensor data for a sister vessel is unavailable, our transfer learning-based approach can provide accurate predictions using noon reports. Sensor data may contain uncertainties due to faulty or poor recordings and high variability \cite{Li2022,Laurie2021}, in which case our approach can be used for data quality management. Noon reports typically have sparse entries, but our method integrates prior knowledge on ship data at high frequency, making it effective in detecting outliers in sensor-based predictions for sister vessels. 
Additionally, adopting our approach to predictive monitoring can help in managing the cost of installing sensors on each sister vessel. 

Our method improved the accuracy of shaft power predictions by 3.6\% for a sister vessel and 5.3\% for a different vessel, compared to a model trained solely on their noon reports. When compared to sensor data models, our approach demonstrated a 2\% performance decrease for a similar vessel and 3\% for a different vessel. Although our approach significantly reduced the predictive gap between sensor data and noon reports, the error compared to sensor data models was higher than that observed for sister vessels. Thus, this approach may not be directly comparable to sensor data models for data quality management, it can still provide more accurate and robust predictions using noon reports when sensor data is unavailable. 

Our transfer-learning-based approach yielded high predictive performance for vessel shaft power predictions comparable to sensor data, even when using a limited amount of noon reports for training each vessel. Our method accurately predicted the power consumption trend for 2024 and 2025. The consistent power predictions over time demonstrated that our approach effectively models the input-output relationship between ship operational data and shaft power, enabling fuel efficiency planning during a voyage and facilitating predictive maintenance. Then, ship operators can predict future performance degradation, even with a limited number of noon reports.

\subsection{Limitations and Future Work}
While our proposed transfer learning-based approach has shown promising results in enhancing shaft power predictions, there are some inherent limitations. Our study is limited by the number of vessels evaluated, as we tested only three sister vessels and one vessel from each of the similar and different categories. Future work should test the approach on a broader range of vessels to address scalability and generalizability issues. For temporal analysis, we trained the base model on data from a specific year and then tested it on the subsequent years. However, it remains unclear how long the model can make accurate predictions without retraining. Investigating the optimal retraining frequency and its long-term performance could provide valuable insights. Additionally, while weather and sea conditions significantly impact shaft power consumption, our approach did not examine their individual effects. Future research should explore which parameters are most effective when transitioning from high-frequency sensor data to low-frequency daily noon reports. The data fusion between publicly available meteorological data and noon reports should be explored. In this study, we utilized sensor data from CMEMS to address weather-related uncertainty, and a similar approach for noon reports may enhance prediction accuracy. The performance of this data fusion strategy should be compared to our transfer learning-based approach to assess differences. Moreover, the model architecture is not optimized to handle large amount of diverse data from various vessel types. The fixed configuration of hidden layers and neurons may not be suitable for all vessels and requires further investigation to improve performance across different scenarios.

In summary, while our approach improves shaft power prediction using noon reports, enhancing scalability, architectural flexibility, and integrating additional data sources are crucial for improving performance across diverse vessels and operational conditions.

\subsection{Data and Code Availability} The proposed transfer learning-based approach was implemented for our industrial partner and therefore it is not made available publicly. Similarly, the data used for the model training and testing is proprietary.

\section{Acknowledgments}
This study has been funded by the Research Council of Norway under grant agreement No. 346603, the GASS project. We thank Global Ocean Waves Analysis and Forecast (\url{https://doi.org/10.48670/moi-00017}) for providing weather-related data used in this work. We thank Joachim Berdal Haga for his contributions to data acquisition and feature extraction, Glenn Terje Lines, Tetyana Kholodna and Adam Sobey for their useful feedback and discussions leading to the shaping of this work.

\bibliographystyle{splncs04}
\bibliography{mybibliography}

\end{document}